\documentclass[preprints,article,accept,moreauthors,pdftex]{Definitions/mdpi}

\firstpage{1} 
\pubvolume{1}
\issuenum{1}
\articlenumber{0}
\pubyear{2024}
\copyrightyear{2024}
\datereceived{ } 
\daterevised{ } 
\dateaccepted{ } 
\datepublished{ } 
\hreflink{https://doi.org/} 
\pdfoutput=1 

\Title{Quantum-Cognitive Tunnelling Neural Networks for Military-Civilian Vehicle Classification and Sentiment Analysis}

\TitleCitation{Quantum-Cognitive Tunnelling Neural Networks for Military-Civilian Vehicle Classification and Sentiment Analysis}

\definecolor{my_green}{rgb}{0.55, 0.71, 0.0}

\usepackage{braket}
\usepackage{comment}
\usepackage{gensymb}

\Author{Milan Maksimovic $^{1}$\orcidA{}, Anna Bohdanets $^{2}$\orcidB{}, Immaculate Motsi-Omoijiade $^{2}$\orcidC{}, Guido Governatori $^{3,1}$\orcidD{} and Ivan~S.~Maksymov $^{1,}$\orcidE{}*}

\AuthorNames{Milan Maksimovic, Anna Bohdanets, Immaculate Motsi-Omoijiade, Guido Governatori and Ivan S.~Maksymov}

\address{%
$^{1}$ \quad Artificial Intelligence and Cyber Futures Institute, Charles Sturt University, Bathurst, NSW 2795, Australia; mmaksimovic@csu.edu.au\\
$^{2}$ \quad Artificial Intelligence and Cyber Futures Institute, Charles Sturt University, Barton, ACT 2600, Australia; abohdanets@csu.edu.au; imotsi@csu.edu.au\\
$^{3}$ \quad School~of~Engineering~and~Technology, Central~Queensland~University, Rockhampton, QLD 4701, Australia; g.governatori@cqu.edu.au}

\corres{Correspondence: imaksymov@csu.edu.au}

\abstract{Prior work has demonstrated that incorporating well-known quantum tunnelling (QT) probability into neural network models effectively captures important nuances of human perception, particularly in the recognition of ambiguous objects and sentiment analysis. In this paper, we employ novel QT-based neural networks and assess their effectiveness in distinguishing customised CIFAR-format images of military and civilian vehicles, as well as sentiment, using a proprietary military-specific vocabulary. We suggest that QT-based models can enhance multimodal AI applications in battlefield scenarios, particularly within human-operated drone warfare contexts, imbuing AI with certain traits of human reasoning.}


\begin{document}

\section{Introduction}
In the evolving landscape of artificial intelligence (AI), quantum cognition theory (QCT) \cite{Khr06, Atm10, Bus12, Pot22} offers a new framework for understanding human perception of the world and decision-making with machine assistance \cite{Bus17, Mak24_APL, Han24, Maks25}. Unlike classical models, QCT uses principles such as superposition, entanglement and interference to explain human reasoning, including conflicting thoughts, context-dependent choices and deviations from classical probability \cite{Bus12, Pot22}.
\begin{figure}
  \centering
  \includegraphics[width=0.999\textwidth]{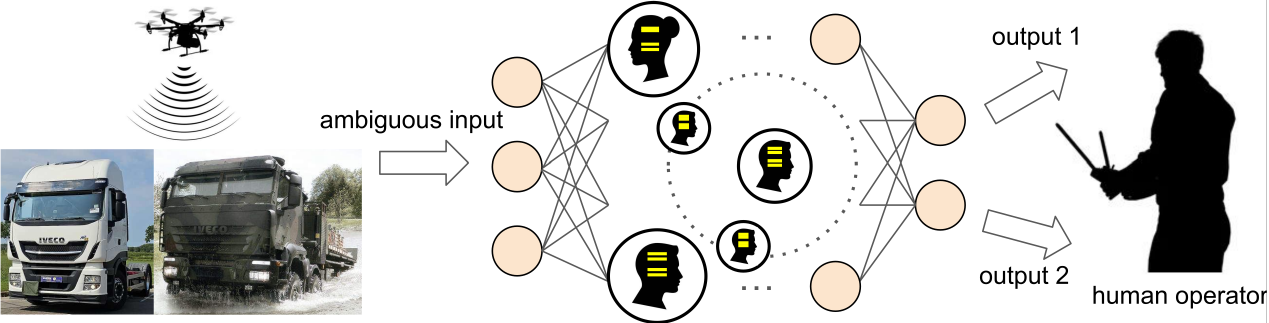}
  \caption{QT enhances machine learning by incorporating models of human-like bistable perception of optical illusions and cognitive biases into the neural network. The principle of energy quantisation aligns with human mental states (depicted by lines on the head silhouettes), where transitions between energy levels enable nuanced military-civilian vehicle differentiation.}
  \label{fig:teaser}
\end{figure}

In the works \cite{Atm10, Bus12}, it was proposed that a quantum oscillator---a fundamental concept of quantum physics \cite{Gri04}---serves as a system demonstrating the ability of QCT to describe human perception more effectively than any existing classical model (e.g., Markov model \cite{Bus12}). Further work \cite{Mak24_illusions} extended this approach by integrating the physical phenomenon of quantum tunnelling (QT) into the oscillator model. The QT approach, which employs the probability of electron transmission through a potential barrier, has been shown to provide a plausible model of both human mental states \cite{Aer22, Aer22_1, Khr23, Mak24_information} and neural mechanisms underlying brain function \cite{Geo18, Geo_book, Man24, Gho25}.

While in classical mechanics a particle confined within a finite region (such as a potential barrier) can have any energy, in quantum mechanics its energy levels are quantised \cite{Gri04}. Mathematically, this property originates from Schr{\"o}dinger’s equation, whose solutions also underpin the effect of QT \cite{Gri04}. Philosophically interpreting quantised energy levels as human mental states (depicted by the human head silhouettes with the lines schematising discrete energy levels---mental states---in Figure~\ref{fig:teaser}), it has been demonstrated that periodic oscillations between two or more energy levels provide a plausible model for optical illusion perception \cite{Mak24_APL}.

The energy level framework also offers insight into a range of interrelated psychological effects and biases exhibited by social groups \cite{Mak24_information} and individuals \cite{Mak24_information1}. This framework can represent both discrete, computer bit-like behaviour---where the states `0' and `1' correspond to being fully certain that a vehicle is military or civilian, respectively---and quantum bit (qubit)-like states, $\ket{0}$ and $\ket{1}$, whose superposition can be interpreted as distinguishing a military vehicle from a civilian one with a certain probability.
\begin{figure}[t]
  \centering
  \includegraphics[width=0.75\linewidth]{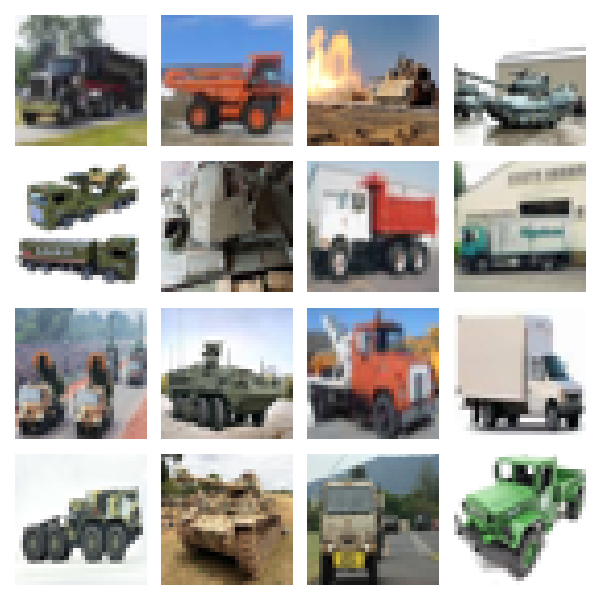}
  \caption{Example of a CIFAR-military vehicle subdataset.}
  \label{fig:images}
\end{figure}

Based on these findings, it has been suggested that AI systems based on QCT can better navigate uncertainty and ambiguity, making them ideal for high-stakes environments such as drone operation \cite{Abb24_review, Maks25}. A feedforward neural network incorporating the effect of QT as a neural activation function was developed and its ability to plausibly reproduce human perception was showcased \cite{Mak24_APL, Maks25}. A mathematical framework attempting to quantify human-like behaviour of the QT model was also proposed \cite{Maks25}.

In this paper, we suggest that QT-based models have the potential to benefit military AI applications by enhancing real-time, adaptive decision-making. Through purpose-built test datasets, we show that QT-based Bayesian and recurrent neural networks \cite{Maks25_1}---which combine classical probabilistic models and memory-based approaches with novel QT techniques---improve accuracy in distinguishing military from civilian objects (see Figure~\ref{fig:teaser} for illustration) and clarifying verbal commands, thus enabling multimodal quantum-inspired AI and holding the potential to minimise civilian casualties by improving the precision of decision-making in complex, high-pressure environments.

\section{Neural network models and test tasks}
A detailed description of QT-based neural network architectures and a comprehensive benchmarking of their performance against their classical counterparts have been conducted in prior work \cite{Mak24_APL, Maks25, Maks25_1, McN25}. Here, we employ a QT-based Bayesian neural network (QT-BNN), where we replaced the standard activation ReLU function with the QT activation function \cite{Maks25_1}. Predictions are made by sampling weights from their distributions, with the final output obtained by averaging multiple samples \cite{Jos22}. The same QT function is used in the recurrent neural network (QT-RNN) \cite{Jun19, Maks25_1}. We note that while the hybridisation of the QT-BNN and QT-RNN models into a standalone multimodal AI solution is beyond the scope of this paper, there are no technical limitations preventing such an approach. This will be explored as part of future work. 

The classification of military and civilian images is a crucial task in training neural network models for military applications \cite{Moy20}. Recent military conflicts have highlighted the growing use of drones in warfare, increasingly targeting vehicles---including civilian ones---driving shifts in combat strategies and operational models \cite{Sur24, Spi25}. The challenge is further compounded by the potential repurposing of civilian vehicles for military operations.

Only a limited number of relevant image classification datasets are publicly available \cite{Gup21}. Despite their usability, their quality remains insufficient for the purposes of this study. To address this gap, we have systematically monitored the types of military and civilian vehicles used the ongoing military conflicts, focusing on trucks, combat tanks, armoured personnel carriers, missile vehicles, armoured cars and military engineering vehicles \cite{Sea22, Spi25}.

Many military vehicles, particularly European-made trucks \cite{Hei20}, share the chassis and general cabin aesthetics with their civilian counterparts (e.g., Iveco, Tatra and Kamaz multirole vehicles; see Figure~\ref{fig:teaser}). We found that civilian trucks used in the ongoing military conflicts also appear in the truck subset of the well-known CIFAR-10 dataset \cite{Kri09}. To extend this dataset, we added 1,860 open-source images of custom-identified military vehicles (Figure~\ref{fig:images}). These images were preprocessed and formatted to be compatible with the CIFAR dataset structure for integration into the training pipeline.
\begin{table}
  \caption{Positive and Negative Words: Military Context}
  \label{tab:military_words}
  \centering
  \begin{tabular}{clcl}
    \toprule
    \multicolumn{2}{c}{\textbf{Positive Words}} & \multicolumn{2}{c}{\textbf{Negative Words}} \\ 
    \midrule
    Achieve       & Advance       & Abort       & Ambiguous \\ 
    Authorize     & Clear         & Breakdown   & Cancel \\ 
    Command       & Confirm       & Compromised & Conflicted \\ 
    Decisive      & Definitive    & Degrade     & Defeat \\ 
    Deploy        & Designated    & Denied      & Disrupt \\ 
    Effective     & Engage        & Doubtful    & Failure \\ 
    Established   & Mission-ready & Ineffective & Misfire \\ 
    Objective-secured & On-target & Obstructed  & Off-course \\ 
    Success       & Validated     & Unconfirmed & Void \\ 
    \bottomrule
  \end{tabular}
\end{table}

To showcase the potential of the proposed QT-based models for use in multimodal solutions, we also created a custom database of terms commonly used in operational planning, rules of engagement and mission status reports and tested in using the QT-RNN model. These words reflect either successful or unsuccessful outcomes when defining and confirming objectives in a military context (Table~\ref{tab:military_words}). They were arranged into meaningful phrases that mimic the communication between the operators of military drones.

While the research presented on the QT-RNN model focuses on textual interpretation and situational analysis, its shared QT-based architecture offers the potential to complement the QT-BNN machine vision model. Although this extension lies beyond the scope of the current study, it represents a promising direction for future research, particularly in enhancing the situational awareness and decision-making accuracy of autonomous defence platforms.

\section{Results and discussion}
\subsection{Machine learning aspects}
\begin{figure}[t]
  \centering
  \includegraphics[width=\linewidth]{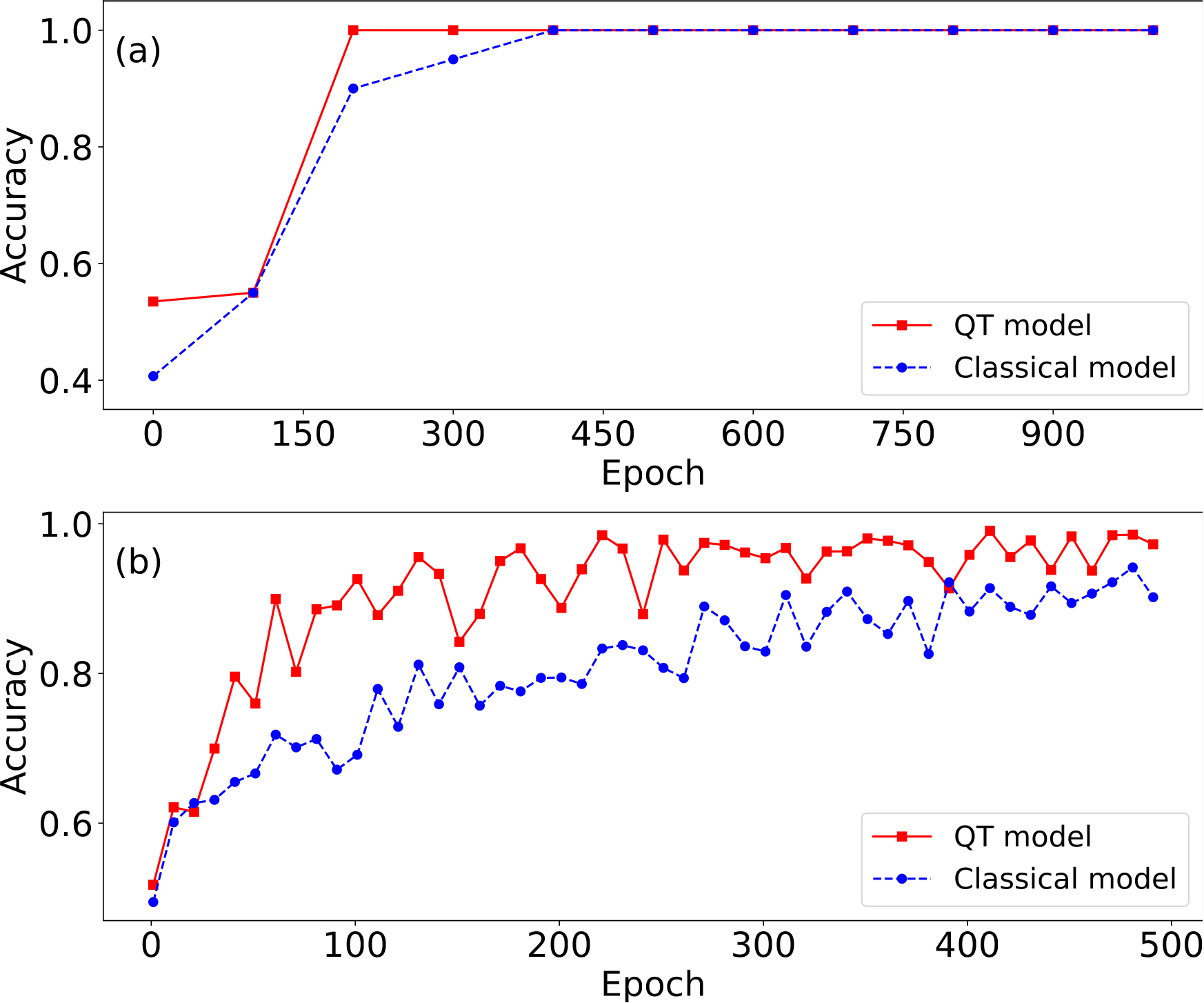}
  \caption{Accuracy over epochs: (a)~QT-RNN sentiment analysis, (b)~QT-BNN military-civilian classification. Accuracy of the respective classical models shown for reference.}
  \label{fig:mil_civil_acc}
\end{figure}

Traditional machine learning procedures often assess model correctness by analysing convergence accuracy versus epochs, ensuring that a model effectively learns patterns from data. As shown in Figure~\ref{fig:mil_civil_acc}(a), the QT-RNN model achieves 100\% accuracy within just 300 epochs, starting at an initial accuracy of 95\%. In contrast, the classical model that used the standard ReLU activation function requires at least 400 epochs to fully converge, exhibiting a significantly slower learning dynamics. This behaviour is characteristic of the QT models and has been previously discussed in great detail in Ref.~\cite{Maks25, Maks25_1}.

Figure~\ref{fig:mil_civil_acc}(b) illustrates the training loss and accuracy of the QT-BNN model as a function of the number of epochs, comparing it with the accuracy of a classical feedforward neural network with the same number of hidden-layer nodes and batch size \cite{Maks25_1}. Outperforming the classical model, the QT-BNN model achieves a maximum training accuracy of 99.06\%, with an average accuracy of approximately 92.13\% after epoch 400.

\subsection{Cognitive modelling aspects}
Thus far, the discussion of the QT-based models has focused on their machine learning aspects, demonstrating their advantage over the traditional classical models. However, one of the main hypotheses of this paper is that QT models can not only be more powerful than traditional neural network architectures, but also introduce certain aspects of human decision-making~\cite{Mak24_APL}.

Prior work using more technically simple QT-based feedforward networks applied to the Fashion MNIST dataset has established a statistical framework that suggests that QT-based models exhibit a degree of human-like behaviour in their predictions \cite{Maks25, Maks25_1}. Specifically, it was shown that while both classical and QT models are not flawless from a machine learning perspective, predictions made by the QT model tend to make more `common sense' (for a more detailed relevant discussion, see Refs.~\cite{Gal24, Mak24_illusions, Mak24_APL, Mak24_information, Mak24_information1}).

Similarly, CIFAR-format images of trucks can often be misclassified to their visual similarities with other vehicles. In the military-civilian vehicle dataset, trucks can be misclassified as military vehicles because of their comparable size and shape \cite{Hei20}. Additionally, fire trucks can be mistaken for regular trucks or even military vehicles if their distinctive features (like ladders, sirens or red colour) are not clearly distinguishable or misinterpreted. Lastly, construction vehicles such as dump trucks or cement mixers, while similar to regular trucks, often have additional features that can lead to confusion.

Psychological aspects should also be included in the discussion, especially because red is associated by humans with danger or aggression \cite{Sol95, Kun15}. It is noteworthy that some training images of military vehicles feature self-propelled artillery or combat tanks in action, producing a distinctive red flame (e.g., the second-to-last image in the top row of Figure~\ref{fig:images}). Consequently, training on these images can imbue the model with a human-like association of red with military danger, resulting in the misclassification of fire trucks and other red-coloured vehicles as military ones.

To verify the above hypothesis, we systematically analysed images of civilian vehicles that were misclassified as military ones by the QT-BNN and classical models. Although we acknowledge that the number of military vehicle images in the test dataset is relatively low compared to the previously used Fashion MNIST dataset \cite{Maks25, Maks25_1}, our findings are consistent with the results of the previous study. While neither model is perfect, the output of the QT-BNN aligns more closely with common logical reasoning.

\begin{figure}[t]
  \centering
  \includegraphics[width=0.75\linewidth]{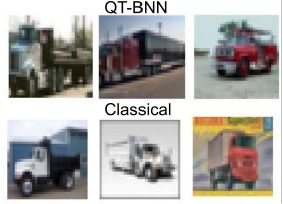}
  \caption{Representative examples of civilian vehicles misclassified as military by the QT-BNN and classical models.}
  \label{fig:confus_mil}
\end{figure}

In fact, as shown in the top row of Figure~\ref{fig:confus_mil}, the civilian truck images misclassified by the QT-BNN closely resemble military vehicles, aligning with the logical arguments presented above. On the contrary, while the overall accuracy of the classical model---defined mathematically---may be comparable to that of the QT-BNN, the misclassifications it produces, as shown in the bottom row of Figure~\ref{fig:confus_mil}, are less interpretable based on common sense. For example, two civilian trucks misclassified as military have white cabins, a feature not typically used in military settings.

\section{Future work}
Thus, both the prior work \cite{Maks25, Maks25_1} and the current paper provide strong evidence supporting the ability of the QT model to mimic human-like perception and judgement in decision-making. The statistical model developed in Ref.~\cite{Maks25} also demonstrated that the QT model classifies images more flexibly than traditional approaches and aligns more closely with human cognitive processes. Comparisons of outputs, weight distributions and uncertainty measures further confirmed that the QT model trains more efficiently and delivers comparable or improved accuracy relative to similarly configured conventional models. Nevertheless, further human trials are required for definitive validation, as discussed below.

QT models have also been calibrated \cite{Mak_arxiv, Mak24_illusions, Mak24_APL} using electroencephalogram \cite{Joo20} and eye-tracking data \cite{Cho20} from humans monitoring and self-reporting optical illusions. Similar methods can be applied to assess decision-making in classifying civilian and military vehicles. This involves software displaying CIFAR-format images to participants, who classify them as 'military' or 'civilian'. The study should address ethical considerations, ensuring informed consent, data handling and demographic data collection. Randomisation of image presentation is necessary to reduce bias, with image duration calibrated to reflect the time constraints faced by military drone operators, who make decisions under pressure \cite{Aza20}.

Subsequent investigations will also involve further calibration of the model output using human-produced experimental data. In the QT model, the hyperparameters include the width and height of the potential barrier, which control the structure of the discrete quantum energy levels \cite{Maks25, Maks25_1}. Since these energy levels have been linked to human mental states \cite{Mak24_information}, theoretical progress can be made by adjusting the quantum behaviour of the model to more closely align with human cognitive and emotional responses.

Future work within the emphasised directions will also advance a three-level ethical approach to military AI, comprising data, performance and values. This situates the proposed model within the inherent logic of responsible military AI \cite{Anna1}, thereby encompassing considerations such as transparency and explainability, privacy and data protection, robustness and safety, and inclusion and anti-discrimination---all of which collectively impact notions of trust, accountability and fairness \cite{Anna2}.

At the data level, the approach to data curation should follow the debiasing and transparency principles, as demonstrated in the selection of the image dataset from open-source and CIFAR-10 databases. The present study has not engaged with sensitive data in the building, training or evaluation of the model. At the performance level, the present approach leads to faster and better results with limited data (purpose-built test datasets), enhancing its practical relevance in resource-constrained environments without compromising quality. Additionally, studies related to performance have shown that systems integrating human and model predictions are more accurate than either humans or the model alone \cite{Anna3}. Further studies on deepfake detection have also shown that humans outperform AI models, owing to traits such as emotional intelligence, conscientiousness and a prevention-focused mindset~\cite{Anna4}.

By integrating both human and machine capabilities into the model-building process, the approach proposed in this work can offer improved explainability. It has the potential to provide insights into the nuances of human decision-making, including judgements considered to be `common sense', which the model seeks to approximate. This, in turn, reinforces human-centricity as a core value in AI development \cite{Anna5}. Moreover, it supports the previously articulated view that differences in machine learning methodologies can influence a model’s ability to align with the principles of responsible AI \cite{Anna6}. Generally speaking, this study renders concerns about fairness and inclusion less prominent, as the model explicitly incorporates elements of human cognition involved in visual recognition---such as pattern recognition, categorisation and abstract reasoning. Importantly, the findings are expected to deepen our understanding of cognition within the context of human–machine teaming, particularly in contemporary military operations.

\section{Conclusions}
Thus, this study demonstrates the potential of quantum tunnelling (QT)-based neural networks---based on quantum cognition theory---to enhance machine learning models tasked with complex, high-stakes classification problems. By integrating QT principles into neural architectures, we demonstrated the ability to mimic certain aspects of human perception relevant to recognising ambiguous inputs and interpreting sentiment, particularly in sensitive military contexts. We evaluated the models on two benchmarking tasks:~distinguishing between military and civilian vehicles using a customised CIFAR-format dataset and performing sentiment analysis with a domain-specific military lexicon.

The exploratory, `blue sky' nature of this study is noteworthy, as it remains largely a work in progress. Readers are encouraged to engage with, adapt and build upon the computational codes and datasets available via the links provided in the Data Availability section.

To ensure accuracy, fairness and contextual relevance, the datasets were curated from publicly available sources, with careful attention to represent real-world vehicle models and language used in active conflict zones. The study also aims to address ethical considerations by promoting transparency in algorithm design and use, minimising risks of bias or misuse and supporting responsible human-AI collaboration in battlefield scenarios such as drone warfare. Overall, our findings support the view that QT-based models can serve as a foundation for robust, human-aligned AI systems capable of flexible reasoning and improved performance in ambiguous, multimodal or adversarial environments.

Finally, although the QT-based neural models examined in this work have been numerically simulated using rigorous theoretical models of quantum tunnelling, our related studies \cite{Maks25_1, McN25} demonstrate the feasibility of implementing these models in hardware, particularly through quantum electronic devices such as tunnel diodes. Specifically, we have demonstrated that analogue circuits employing tunnel diodes can implement the QT-based neural network activation functions, which in this work have been shown to outperform traditional activation functions in both standard classification tasks and cognitively motivated problems \cite{Maks25_1, McN25}. Moreover, the suggested hardware implementation of the QT-based neural models promises certain advantages over traditional quantum computing-based counterparts, even those capable of operating at room temperatures \cite{QuantumBrilliance2025}. Indeed, existing commercial quantum systems remain expensive, fragile and relatively power-hungry, making them impractical for deployment on large fleets of drones. Given that modern warfare increasingly treats drones as expendable assets \cite{Abb24_review}, the efficiency and robustness of QT-based architectures could offer a significant operational advantage.

\vspace{6pt} 

\authorcontributions{The idea for this paper was conceived jointly by all authors. M.M. and I.S.M. developed the computational framework and created the test datasets. The machine learning performance aspects were discussed with input from G.G. A.B. and I.M.O. led the ethics of AI component of the research, with contributions from G.G. All authors contributed to the discussion and editing of the final manuscript and approved it for submission.}

\funding{This research received no external funding}

\institutionalreview{Not applicable.}

\informedconsent{Not applicable.}

\dataavailability{The source codes that implement the neural network models discussed in this paper are available in the GitHub repository, \url{https://github.com/IvanMaksymov/Quantum-Tunnelling-Neural-Networks-Tutorial} and \url{https://github.com/IvanMaksymov/QT-CIFAR-Military}.}


\conflictsofinterest{The authors declare no conflicts of interest.} 



\abbreviations{Abbreviations}{
The following abbreviations are used in this manuscript:\\

\noindent 
\begin{tabular}{@{}ll}
AI & artificial intelligence\\
CIFAR & The Canadian Institute for Advanced Research\\
MNIST & Modified National Institute of Standards and Technology database\\
QT-BNN & quantum-tunnelling Bayesian neural network\\
QT-RNN & quantum-tunnelling recurrent neural network\\
QCT & quantum cognition theory\\
QT & quantum tunnelling\\
ReLU & rectified linear unit\\
\end{tabular}
}

\begin{adjustwidth}{-\extralength}{0cm}

\reftitle{References}


\bibliography{refs}

\end{adjustwidth}
\end{document}